# Two Heads Are Better Than One: Averaging along Fine-Tuning to Improve Targeted Transferability

Hui Zeng†
Southwest University of Sci. & Tech.
Mianyang, China
zengh5@mail2.sysu.edu.cn

Sanshuai Cui†
City University of Macau
Macau, China
sanshuaicui@cityu.edu.mo

Biwei Chen*
Beijing Normal University
Zhuhai, China
bchen@bnu.edu.cn

Anjie Peng
Southwest University of Sci. & Tech.
Mianyang, China
penganjie200012@163.com

*Abstract*—With much longer optimization time than that of untargeted attacks notwithstanding, the transferability of targeted attacks is still far from satisfactory. Recent studies reveal that fine-tuning an existing adversarial example (AE) in feature space can efficiently boost its targeted transferability. However, existing fine-tuning schemes only utilize the endpoint and ignore the valuable information in the fine-tuning trajectory. Noting that the vanilla fine-tuning trajectory tends to oscillate around the periphery of a flat region of the loss surface, we propose averaging over the fine-tuning trajectory to pull the crafted AE towards a more centered region. We compare the proposed method with existing fine-tuning schemes by integrating them with state-of-the-art targeted attacks in various attacking scenarios. Experimental results uphold the superiority of the proposed method in boosting targeted transferability. The code is available at *github.com/zengh5/Avg_FT*.

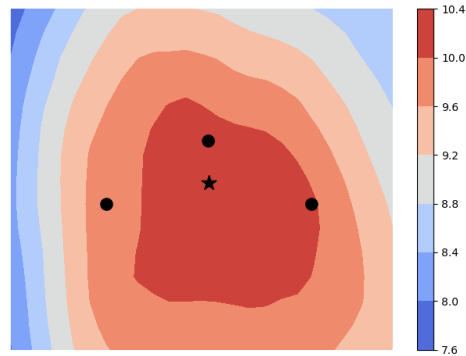

Fig. 1. Conceptual illustration of averaging along fine-tuning. The circles are snapshots along the fine-tuning trajectory, and the star denotes the barycentre of the triangle determined by three circles. The AEs are crafted against a pretrained Resnet50 model, and the loss surface is calculated with a pretrained Resnet18.

## I. Introduction

Building an interpretable and trustworthy deep neural network (DNN) is critical in many security-sensitive applications. Adversarial example (AE) is a valuable tool for uncovering the opaqueness and vulnerability of DNNs [1, 2]. In a broad sense, AE generation optimizing pixel values against DNN models is analogous to DNN training optimizing model weights guided by (data, label) pairs. Consequently, improving AEs' transferability across various models is akin to enhancing DNNs' generalizability to unseen data. Many tricks for enhancing model generalization in machine learning have been applied to improve AE transferability. For example, the widely accepted momentum method [3] has been used to stabilize the AE optimization [4, 5], and input transformation-based attacks [5-7] can find their consistency-enforcing [8] counterparts in machine learning.

This paper centers on the transferability of targeted attacks, which is much less studied yet more daunting than their untargeted counterparts. Previous studies have shown that feature-space fine-tuning is an efficient and effective way of boosting targeted transferability [9-11]. However, these methods only utilize the endpoint of the fine-tuning and fail to exploit the trajectory information fully. Inspired by the fact that averaging high-quality model weights leads to better generalization [12, 13], we propose averaging AEs along the fine-tuning trajectory to generate more transferable AEs. We hypothesize and empirically validate that by temporal averaging, the crafted AE is calibrated towards a more centered region of the loss surface (Fig. 1), thus exhibiting stronger transferability. This paper substantially improves [11] in the following aspects:

- We discover that AEs obtained by [11] typically oscillate around the periphery of a flat region of the loss surface.

- Based on this observation, we propose a novel targeted attack, averaging along fine-tuning (AaF), where crafted AEs are located in a more central area than that in [11], thereby exhibiting stronger transferability across different DNNs.

- Experiments demonstrate that AaF can improve state-of-the-art targeted attacks notably with negligible overheads.

## II. Threat Model and Related Work

### A. Threat model

We define our threat model from three perspectives: the goal, the knowledge, and the capability of the adversary.

**Adversary's goal.** This paper copes with the targeted attack, in which an adversary aims to mislead a DNN-based classifier $f()$ to predict an AE $I'$ as a specific label $y_t$, i.e., $f(I') = y_t$.

**Adversary's knowledge.** We address the transfer-based scenario in which an adversary does not know victim models.

*Corresponding author. †Equally contributed. This work was supported by the Opening Project of Guangdong Province Key Laboratory of Information Security Technology (no. 2023B1212060026) and the Start-up Scientific Research Project for Introducing Talents of Beijing Normal University at Zhuhai (no. 312200502504).

What he can do is resort to a surrogate model for crafting AEs.

**Adversary's capability.** Under human inspection, well-crafted AEs should be non-suspicious. Without loss of generalizability, we constrain the perturbation budget with the $L_\infty$ norm in this paper. To improve the attack power, resource-intensive schemes train extra, target-specific classifiers [14] or generators [15-17]. However, the training time will be prohibitive when the number of targeted classes is enormous. Hence, we follow the conventional simple iterative attacks that require neither additional data nor model training in this study.

*B. Transferable targeted attack*

A targeted attack is strictly more challenging than an untargeted one since the latter can be regarded as targeted towards the easiest label [18]. Moreover, unique challenges prevail in targeted attacks, e.g., gradient vanishing [19, 20] and the restoring effect [19, 21], which necessitates tailored schemes to craft transferable targeted AEs.

To address the gradient vanishing challenge, the Po+Trip attack [19] and the Logit attack [20] replace traditional cross-entropy (CE) loss with the Poincare distance loss and Logit loss, respectively. More recently, the Margin attack [22] argues that downscaling the logits with a temperature factor or an adaptive margin can also alleviate the issue caused by gradient vanishing. To overcome the restoring effect, [19] introduces a triplet loss to push $I'$ away from $y_o$, and the SupHigh attack (SH) further suppresses all high-confidence labels [21]. In addition, the self-universality method (SU) introduces a feature similarity loss to encourage the adversarial perturbation to be self-universal [23]. The activation attack (AA) [24] pushes $I'$ towards a carefully selected natural image of the target class in the feature space. The clean feature mixup method (CFM) mixes features from other images in the same batch to encourage competition during optimization [25]. Strictly speaking, AA and CFM go beyond simple gradient methods since additional images are involved in attacking.

Another interesting research direction is fine-tuning an existing AE in the feature space. The intermediate level attack (ILA) [9] maximizes the scalar projection of the AE on the direction $f_l(I') - f_l(I)$, where $f_l()$ is the feature presentation in the $l$-th layer. On the other hand, the feature-space fine-tuning method (FFT) [11] is guided by a combined aggregate gradient:

$$\overline{\Delta}^{combine} = \overline{\Delta}^{I',t} - \beta \overline{\Delta}^{I,o} \quad (1)$$

where $\overline{\Delta}^{\cdot,t}$ is the aggregate gradient measuring the $y_t$-related feature importance; $\overline{\Delta}^{\cdot,o}$ measures the $y_o$-related feature importance. Combining them together with a predefined weight $\beta$ could encourage the features contributing to $y_t$ and suppress $y_o$-related features simultaneously. Because targeted attacks demand significantly more iterations to converge than untargeted ones [20], fine-tuning is a promising strategy for enhancing targeted transferability in practice.

Resource-intensive attacks have reported competitive targeted transferability. In the feature distribution attack [14], a one-versus-all classifier is trained for each $y_t$ at each specific layer to predict the probability that a feature map is from $y_t$. The transferable targeted perturbation (TTP) [15] trains class-specific generators to synthesize targeted perturbation. However, training a dedicated generator for every (*source model*, *target class*) pair is costly in practice. This limitation is partially lifted by training a conditional generator [26] to target multi-class simultaneously [16, 17]. However, the number of targeted labels that a single generator can cover is limited, and as the number of targets increases, the attack ability inevitably decreases.

III. PROPOSED SCHEME

*A. Motivation*

Prior research points out that AE generation can be regarded as a dual optimization problem of DNN training [4, 5]. Following this line, AEs' transferability corresponds to DNNs' generalizability. For example, model ensemble, analogous to data augmentation in DNN training, has been proven to effectively enhance attack transferability [27]. However, an ensemble of substitute models may not be feasible in practice. Hence the question: *Can we elevate targeted transferability with a single substitute model*?

The answer to the question above is analogous to improving DNNs' generalizability without training data augmentation. In [12, 13], the authors find that averaging high-performing snapshots along the training trajectory results in a more generalizable DNN. Similarly, during the fine-tuning process of FFT [11], many 'high-quality' snapshots of AEs show muscular attack strength against the surrogate model. Fully utilizing fine-tuning trajectory information rather than just the endpoint inspires the proposed method.

*B. Averaging along fine-tuning*

This paper follows the fine-tuning framework in [11] but attempts to explore better the knowledge contained in the fine-tuning trajectory, as displayed in Fig. 2. Similar to FFT, we start from an AE $I'$ generated by a baseline attack (e.g., CE or Logit). The first step is calculating the feature importance to guide the subsequent fine-tuning. Here, we adopt the aggregate gradient to measure the importance of the feature. Let $\Delta_k^{I',t} = \frac{\partial l_t(I')}{\partial f_k(I')}$, where $l_t()$ denotes the logit output *w. r. t.* $y_t$, The aggregate gradient $\overline{\Delta}_k^{I',t}$ is calculated as:

$$\overline{\Delta}_k^{I',t} = G(\frac{1}{C} \sum_n \Delta_k^{I' \odot M^n, t}) \quad (2)$$

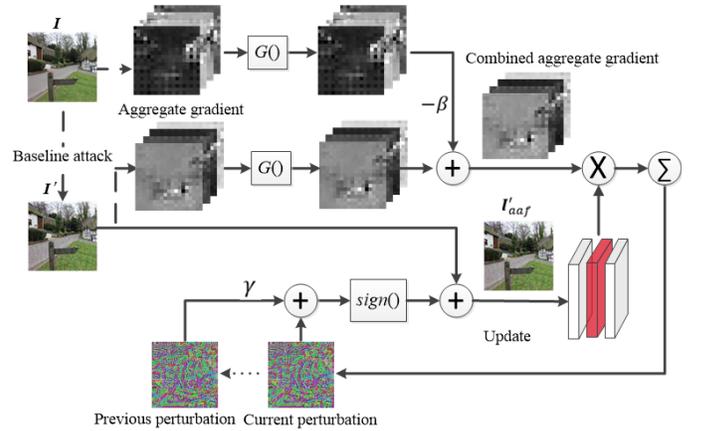

Fig. 2. Overview of the proposed averaging along fine-tuning scheme.

where $C = ||\sum_n \Delta_k^{I',\odot M^n,t}||_2$ is a normalization term, $M^n$ is a random mask to neutralize model-specific information. Unlike FFT, which adopts a pixel-wise mask, in this paper, we employ a patch-wise mask as proposed in [28]. Motivated by [29] that CNN models' discriminative regions are often clustered, we apply Gaussian smoothing $G()$ to the aggregate gradients to further highlight the class-related features. Similarly, $y_o$-related feature importance is calculated as

$$\bar{\Delta}_k^{I,o} = G(\frac{1}{C}\sum_n \Delta_k^{I\odot M^n,o}) \quad (3)$$

Following [11], the aggregate gradients $\bar{\Delta}_k^{I',t}$ and $\bar{\Delta}_k^{I,o}$ are combined, as in (1), to push the subsequent optimization closer to $y_t$ and away from $y_o$.

In [11], $I'$ is optimized according to the following objective:

$$\underset{I'_{ft}}{argmax} \sum (\bar{\Delta}_k^{combine} \cdot f(I'_{ft})), \quad s.t. ||I'_{ft} - I||_\infty \leq \epsilon \quad (4)$$

In contrast, this paper utilizes the whole trajectory of fine-tuning, as illustrated in Fig. 1, and it can be realized as:

$$I'_{aaf} = \gamma^{N_{ft}-1} I'_{ft,0} + \gamma^{N_{ft}-2} I'_{ft,1} + \cdots + I'_{ft,N_{ft}-1}, \quad (5)$$

where $I'_{ft,i}$ represents the $i$-th snapshot during fine-tuning, $N_{ft}$ is the number of fine-tuning iterations, and $\gamma$ is the decaying factor. Eq. (5) reduces to a simple average when $\gamma = 1$, and it means only the endpoint is kept when $\gamma = 0$, as done in [11]. Eq. (5) can be memory-friendly implemented as:

$$I'_{aaf,0} = I'_{ft,0},$$
$$I'_{aaf,i} = \gamma I'_{aaf,i-1} + I'_{ft,i} \quad (6)$$

In the remaining, we abbreviate the proposed fine-tuning scheme (5) as Average along Fine-tuning (AaF).

Recall that the snapshots involved in (5) are expected to possess 'high-quality.' To ensure that, warming-up fine-tuning iterations are adopted before AaF. We find that just a few iterations of warming up ($N_{wu}$) are sufficient to ensure the quality of the snapshots and set $N_{wu} = 5$ in our experiments.

*C. What is AaF doing?*

AaF is based on averaging high-quality AEs along the fine-tuning trajectory. We expect the averaging to encourage the AE generation towards a more centered region than the plain fine-tuning. To verify our conjecture, we plot the logit values (w.r.t. $y_t$) in a plane defined by AEs before fine-tuning ($I'$), fine-tuned with FFT ($I'_{FFT}$) and AaF ($I'_{AaF}$) in Fig. 3. The baseline attack is a Logit attack against a Resnet50 model (Res50) [30], and the logit values are based on an ensemble of pretrained models: Inceptionv3 (Inc-v3) [31], DenseNet121 (Dense121) [32], and VGG16bn (VGG16) [33]. While both FFT and AaF help the AE generation towards the region of higher logit values, AEs crafted by AaF are located in a more central and flatter area.

IV. EXPERIMENTAL RESULTS

We compare the proposed AaF method against two fine-tuning strategies: the targeted version of ILA [9] and FFT [11]. Five representative simple iterative targeted attacks, CE [34], Logit [20], Margin [22], SH [21], and SU [23], are used to craft the baseline AEs for fine-tuning. The Po+Trip attack [19] is omitted because more recent methods claim domination over it. We also investigate how simple iterative attacks with fine-tuning can approach generative methods: TTP [15] and C-GSP [16]. All the iterative schemes start with the TMDI attack [4, 6, 7]. More experimental results are available in our supplementary material: *github.com/zengh5/Avg_FT/supp.pdf*.

*A. Experimental settings*

**Dataset.** Our experiments are conducted on the ImageNet-compatible dataset comprised of 1000 images [35]. All these images are cropped to $299 \times 299$ pixels before use.

**Networks.** Following [20, 22], we use four pretrained models of diverse architectures: Inc-v3, Res50, Dense121, and VGG16 as the surrogates. The AEs' transferability is evaluated on models that have not been used as surrogates. In Sec. IV. C, an adversarially trained model Res50adv serves as the surrogate. We also test four transformer-based models, Swin [36], vit_b_16 [37], pit_b_24 [38], and visformer [39], in the supplementary file.

**Parameters.** For all competitors, the maximum perturbation is set to $\epsilon \leq 16$, and the step size is 2. Following [11], we set the baseline iteration number $N$=160 ($N$=200 when fine-tuning is disabled), the fine-tuning iterations $N_{ft}$=10, and the balance weight $\beta$=0.2. For the fine-tuning layer $k$, we select *Mixed_6b* for Inc-v3, *Conv4_3* for VGG16, and the last layer of the third block for Res50 and Dense121. The decaying factor $\gamma$ in (5) is experimentally set to 0.8. The ablation study on $\gamma$ is provided in the supplementary file. Throughout this section, the best results are in **bold**.

*B. Normal surrogates*

Table I reports the targeted success rates (random-target) for various (surrogate, victim) pairs. All the baseline attacks benefit from fine-tuning, and the improvement led by the proposed AaF is more significant than that of ILA and FFT. As a rule of thumb, the weaker the baseline, the more significant the improvement. Hence, the upturn is particularly salient for the CE attack. For example, when transferring from Dense121 to VGG16, the success rate of the CE+AaF attack is four times higher than that of the CE attack (50.3% vs. 11.3%). Another takeaway is that the more challenging the transfer scenario, the more significant the improvement brought out by fine-tuning. Let Logit be the baseline attack, its gain from AaF is about ten percent in the case of 'Res50→Dense121', whereas in the case of 'Res50→Inc-v3', the success rate is almost doubled. As in [11, 20], we also conduct a worst-case transfer experiment in which the target labels are attached to the least likely classes. The corresponding results are provided in the supplementary file due to page limitations. The brief conclusion is that the improvement from fine-tuning under the most difficult-target scenario is more remarkable than that under the random-target scenario.

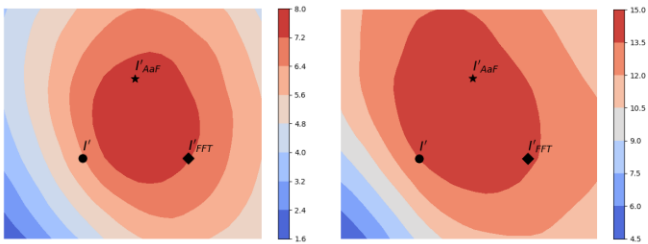

Fig. 3. AaF leads to AEs closer to the center of high transferability. The AEs are crafted against a pretrained Res50 model, and the contours are plotted according to the logit values w.r.t. the target label based on an ensemble of target models. More examples are available in our supplementary file.

TABLE I. TARGETED TRANSFER SUCCESS RATE (%): NO FINE-TUNING/ FINE-TUNING (ILA/FFT/AaF), IN THE RANDOM-TARGET SCENARIO.

| Attack | Source Model: Res50 | | | Source Model: Dense121 | | |
|---|---|---|---|---|---|---|
| | →Inc-v3 | →Dense121 | →VGG16 | →Inc-v3 | →Res50 | →VGG16 |
| CE | 3.9/9.8/9.0/**13.6** | 44.9/61.9/60.4/**66.3** | 30.5/41.1/49.3/**60.2** | 2.3/11.8/13.2/**16.2** | 19.0/36.2/45.3/**50.7** | 11.3/27.6/34.8/**50.3** |
| Logit | 9.1/15.7/15.8/**18.7** | 70.0/73.5/75.3/**78.6** | 61.9/63.6/64.1/**69.3** | 7.8/14.0/15.1/**16.8** | 42.6/53.2/56.7/**60.1** | 37.1/49.7/49.3/**53.2** |
| Margin | 11.0/17.7/14.6/**20.4** | 71.0/74.1/74.5/**78.3** | 61.0/68.4/69.3/**74.7** | 7.5/**19.9**/19.6/19.3 | 45.1/60.0/62.2/**64.5** | 33.5/54.3/54.4/**56.6** |
| SH | 9.6/16.9/16.3/**20.2** | 74.3/77.1/75.7/**79.5** | 63.5/**71.3**/69.8/70.2 | 8.7/15.1/15.9/**17.9** | 48.1/58.8/61.6/**65.2** | 40.5/53.2/52.2/**61.6** |
| SU | 11.1/16.5/18.6/**19.4** | 72.5/72.8/73.3/**77.7** | 64.7/**70.4**/68.5/70.1 | 10.0/17.2/17.1/**19.7** | 49.2/61.5/62.3/**63.3** | 43.1/54.7/55.2/**57.8** |
| Attack | Source Model: VGG16 | | | Source Model: Inc-v3 | | |
| | →Inc-v3 | →Res50 | →Dense121 | →Res50 | →Dense121 | →VGG16 |
| CE | 0.0/0.0/0.0/**0.6** | 0.3/1.9/2.8/**5.9** | 0.5/3.9/3.3/**6.1** | 1.8/2.5/4.7/**7.1** | 2.1/4.3/7.8/**11.5** | 1.5/2.4/4.0/**7.2** |
| Logit | 0.8/1.0/1.1/**2.1** | 10.2/14.1/15.5/**16.9** | 11.6/12.8/15.1/**15.5** | 2.4/4.7/6.3/**9.0** | 4.1/7.9/10.2/**13.2** | 2.2/5.8/7.7/**11.8** |
| Margin | 0.6/0.8/1.0/**1.3** | 7.5/12.7/15.1/**17.6** | 12.0/15.6/18.5/**19.4** | 1.0/3.9/3.6/**6.6** | 3.0/5.6/8.2/**9.5** | 1.9/5.4/8.5/**9.1** |
| SH | 0.8/1.3/**2.7**/2.2 | 11.2/14.4/16.2/**18.5** | 13.6/14.7/**19.9**/19.3 | 2.3/3.7/5.9/**8.1** | 4.5/7.1/10.3/**12.4** | 2.2/5.8/9.6/**11.5** |
| SU | 0.9/1.5/1.7/**1.9** | 13.7/16.4/16.9/**17.8** | 15.7/18.6/19.3/**19.8** | 3.0/5.1/6.3/**8.5** | 4.6/6.9/11.2/**11.8** | 3.5/4.3/6.7/**10.2** |

TABLE II. TARGETED TRANSFER SUCCESS RATE (%): NO FINE-TUNING/ FINE-TUNING (ILA/FFT/AaF). THE AES ARE CRAFTED AGAINST RES50ADV.

| Attack | Source Model: Res50adv | | |
|---|---|---|---|
| | →Inc-v3 | →Dense121 | →VGG16 |
| CE | 14.4/21.6/21.5/**25.3** | 59.0/62.3/69.5/**72.3** | 24.8/40.3/47.2/**53.9** |
| Logit | 26.1/31.8/32.0/**34.2** | 78.6/**79.5**/76.0/79.3 | 55.9/61.8/61.5/**66.0** |
| Margin | 26.8/33.0/32.1/**38.2** | 82.3/81.8/79.4/**83.7** | 55.6/63.4/65.1/**69.2** |
| SH | 21.4/30.5/30.8/**33.4** | 80.7/**81.8**/79.5/80.4 | 67.8/72.2/71.9/**74.1** |
| SU | 27.6/32.7/31.6/**34.3** | 79.9/79.8/75.8/**80.3** | 56.8/64.8/62.3/**66.7** |

TABLE III. TARGETED TRANSFER SUCCESS RATE (%) ($\epsilon = 8/16$) OF ITERATIVE ATTACKS (WITH AaF) VS. GENERATIVE ATTACKS, AVERAGED OVER 10 TARGETS. SOURCE MODEL: RES50.

| Attack | →Inc-v3 | →Dense121 | →VGG16 | Average |
|---|---|---|---|---|
| Logit+*AaF* | 2.7/23.2 | 44.8/78.8 | 44.4/75.2 | 30.6/59.1 |
| Margin+*AaF* | 2.8/22.6 | 44.1/79.8 | 44.9/76.4 | 30.6/59.6 |
| SH+*AaF* | 3.7/24.4 | **48.3**/**80.5** | **47.1**/77.7 | **33.0**/60.9 |
| SU+*AaF* | 3.9/23.3 | 45.1/79.3 | 45.5/75.7 | 31.5/59.4 |
| TTP | **5.7**/**39.8** | 38.6/79.5 | 44.2/75.4 | 29.5/**64.9** |
| C-GSP | 0.4/30.1 | 8.3/67.5 | 10.2/57.0 | 6.3/51.5 |

*C. Robust surrogate*

Next, we craft AEs with an adversarially trained model Res50adv (the $L_2$ budget of the AEs for training the model is 0.01), which is believed to be conducive to transferability [40]. The victim models are the same as those in the last section except for Res50. Table II presents the targeted transferability in this scenario. AEs crafted against a slightly robust model indeed show stronger transferability in most cases. Nevertheless, AaF can further boost their attack ability by a clear margin. Taking 'Res50adv→Inc-v3' as an example, Logit+AaF surpasses Logit by 31% (34.2% vs 26.1%).

*D. Iterative vs. generative attacks*

Last, we compare AaF+simple iterative attacks with two generative attacks: TTP and C-GSP. Since training dedicated generators for each target label and each source model is prohibitive for our used ImageNet-compatible dataset, we download ten author-released generators and follow the '*10-Targets*' setting of [15]. These generators are trained with Res50 being the discriminator. For C-GSP, we train a 10-target conditional generator with Res50 being the discriminator on the ImageNet '*train*' dataset.

We evaluate all competitors under two perturbation budgets: $\epsilon = 8$ and $\epsilon = 16$. Table III shows that fine-tuned iterative attacks yield comparable or even better transferability (when Dense121 or VGG16 is the victim model) than generative methods. Because the generative methods heavily hinge on semantic patterns [20] (refer to the supplementary file for examples), their performance significantly deteriorates as the budget decreases. In contrast, the iterative methods are more resilient when the budget is low. Taking 'Res50→Dense121' for example, Logit+AaF is equivalent to TTP at $\epsilon = 16$, but has a significant advantage at $\epsilon = 8$ (44.8% vs 38.6%).

V. CONCLUSION

Learning from others' strengths is an ancient wisdom. In this paper, we hypothesize and empirically validate that AEs' transferability can be advanced by fully exploiting fine-tuning trajectories. Specifically, we propose a novel targeted attack called Averaging along Fine-tuning (AaF) to encourage AE generation towards flatter regions than the vanilla feature-space fine-tuning. The superiority of the proposed AaF is validated by integrating it with state-of-the-art iterative targeted attacks in various transfer scenarios. Experimental results corroborate that our AaF is superior to existing fine-tuning schemes and can boost targeted transferability universally, while its overhead compared to baseline attack is negligible.

# Two Heads Are Better Than One: Averaging along Fine-Tuning to Improve Targeted Transferability: supplementary material

*Hui Zeng, Sanshuai Cui, Biwei Chen, and Anjie Peng*

The supplementary document consists of six parts of content: A) The pseudo-code of the proposed method; B) Ablation study on the decaying factor $\gamma$; C) Visualization of FFT and AaF in a 2D subspace; D) Attack performance against transformer-based models; E) Attack performance in the most difficult-target scenario; F) Visual comparison.

## A Pseudo code the AaF attack

Due to page limitation, we provide the pseudo-code of the proposed AaF attack in **Algorithm 1**.

## B Ablation study on $\gamma$

We make an ablation study on the newly introduced decaying factor $\gamma$ of AaF in the random-target scenario. The source models are Inc-V3, Res50, Dense121, and VGG16, the same as our paper. The targeted success rates are averaged over three hold-out models and a VIT-based model, Swin. We let $\gamma$ vary from 0 to 1 with a step of 0.1. Note $\gamma = 0$ reduces to the vanilla FFT method with $N_{ft} = 15$, and $\gamma = 1$ means a simple average over the fine-tuning trajectory. Fig. 1 shows that the optimal $\gamma$ for different source models vary. Generally, $\gamma \in [0.4, 0.8]$ will be a good choice for all surrogates. In our study, we experimentally set $\gamma = 0.8$ for simplicity.

---

**Algorithm 1** AaF attack

---

**Input**: A benign image $I$ with original label $y_o$; target label $y_t$.
**Parameter**: Iterations $N$ of the baseline attack; Fine-tuning iterations $N_{ft}$; decaying factor $\gamma$.
**Output**: Adversarial image $I'_{aaf}$.
1: Mount a baseline attack for $N$ iterations, and obtain an AE $I'$.
2: Calculate aggregate gradient $\bar{\Delta}_k^{I',t}$ from $I'$, and $\bar{\Delta}_k^{I,o}$ from $I$.
3: Obtain the combined aggregate gradient $\bar{\Delta}_k^{combine}$ as (1).
4: $I'_{aaf,0} = I'$
5: **for** $t = 1$ to $N_{ft}$ **do**
6:     Fine-tune $I'_{aaf,t-1}$ with the optimization objective defined in (4) and obtain $I'_{ft,t}$.
7:     $I'_{aaf,t} = \gamma I'_{aaf,t-1} + I'_{ft,t}$
8: **end for**
9: **return** $I'_{aaf,N_{ft}}$.

---

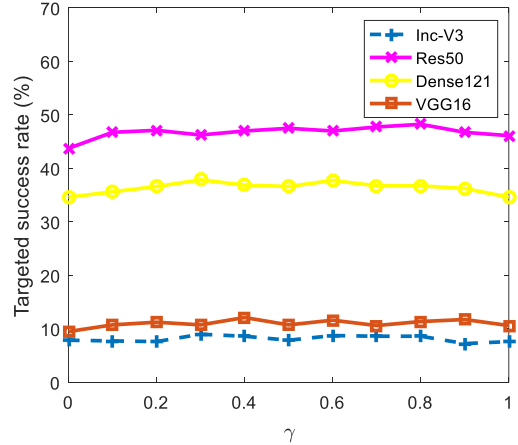

**Fig. 1.** Effect of $\gamma$ on AEs' transferability averaged over three hold-out models and Swin. Each curve represents a different surrogate. The baseline attack is Logit.

## C Visualization of FFT and AaF in a subspace

Besides the examples in the Fig. 3 of the paper, we provide additional examples here. To avoid the bias introduced by cherry pick, we investigate the first 20 samples of the ImageNet-compatible dataset. The planes are generated using the following steps: First, we use the AEs $I'$, $I'_{FFT}$ and $I'_{AaF}$ to span a 2D subspace. Then we calculate the logit of a point $I'_{sample}$ in the spanned subspace based on an ensemble of models:

$$logit_{ensemble} = 1/N (\sum_{i}^{N} logit_i(I'_{sample})),$$

where $N$ is the model number and $logit_i()$ is the logit output w.r.t. the target class of the $i$-th model. The value of $logit_{ensemble}$ indicates the targeted transferability of $I'_{sample}$. As shown in Fig. 2, in most cases, the proposed AaF method steers AE towards a more central region than FFT.

## D Attack performance on transformers

Table 1 reports the targeted transferability against four transformer-based models, Swin, vit_b_16, pit_b_24, and visformer, in the random-target scenario. Compared to the results reported in Table I of the paper, transferring from CNNs to transformers is much more challenging than transferring between CNNs. Nevertheless, in most cases, the proposed AaF is superior to existing fine-tuning schemes. For example, when Res50 is the surrogate, the average success rate of CE+AaF surpasses CE+FFT (the second best) by 46.8% (6.9% vs. 4.7%).

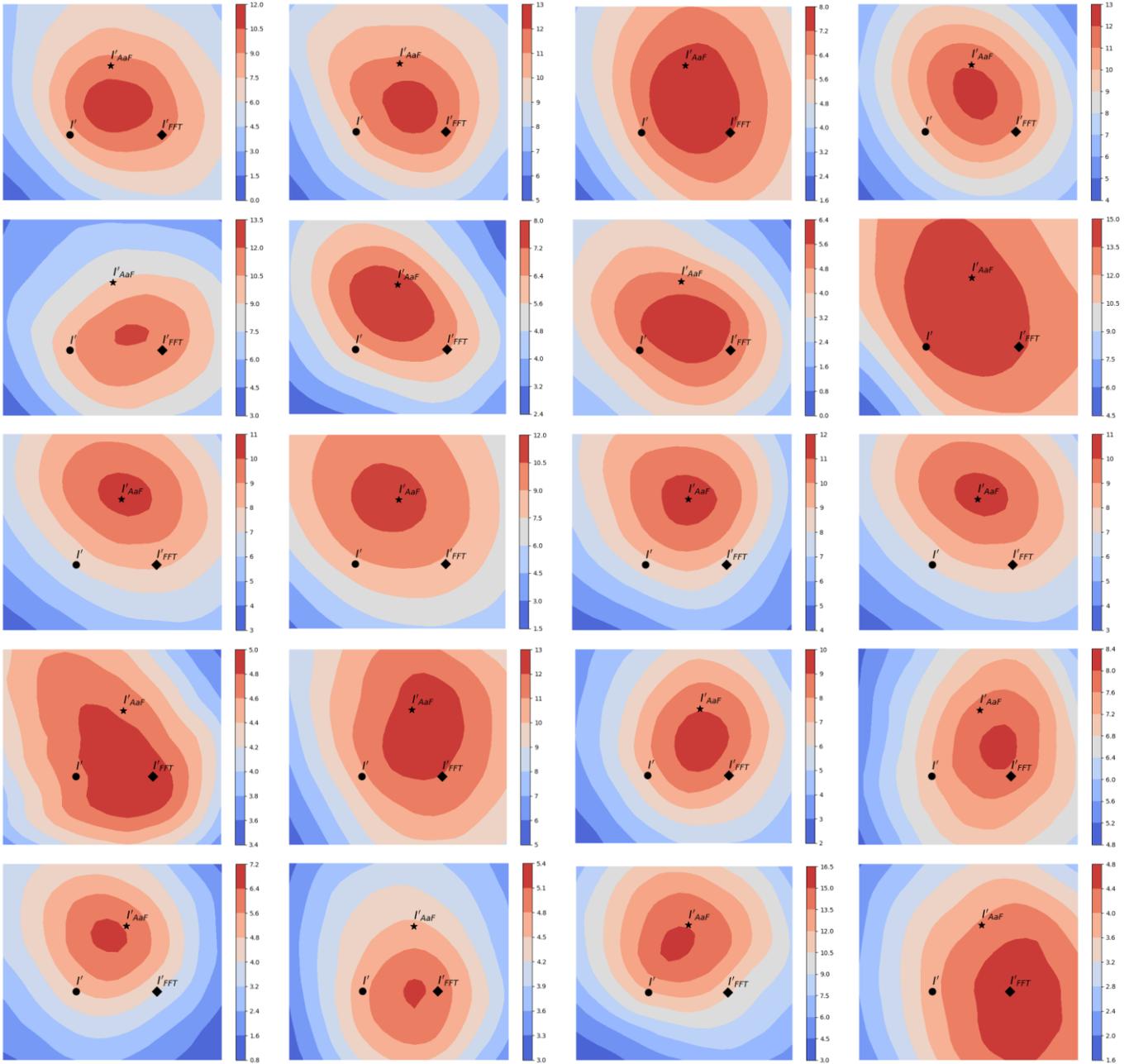

**Fig. 2.** The logit plane of AEs with different fine-tuning schemes. The baseline attack is Logit. The AEs are crafted against a Res50 model, and the logits are calculated based on an ensemble of models: IncV3, Dense121, and VGG16.

### E Attack to the most difficult target

Table 2 compares different fine-tuning schemes in the most difficult-target scenarios. Compared to Table I of the paper, the improvement from fine-tuning is more remarkable under the most challenging target scenario. Let Res50 be the surrogate,

Logit+AaF improves the Logit attack by 18.2% (55.5% vs. 47.0%) in the random-target scenario and 24.2% (38.8% vs. 31.3%) in the most difficult-target scenario, in terms of targeted transfer rate averaged over three victim models.

### F Visual comparison

We visually compare AEs crafted with different methods in Fig. 3. The target label for all samples is 'hippopotamus.' The perturbations introduced by TTP are more suspicious under human inspection, while those introduced by the iterative methods resemble noise.

**Table 1.** Targeted transfer success rate (%) against transformers in the random-target scenario. No fine-tuning/ fine-tuning (ILA/FFT/AaF).

| Source Model | Attack | Victim Model | | | | AVG |
|---|---|---|---|---|---|---|
| | | Swin | vit_b_16 | pit_b_24 | visformer | |
| Res50 | CE | 5.1/7.5/7.7/**11.5** | 0.6/0.8/1.8/**2.7** | 2.0/2.2/2.2/**3.2** | 4.8/8.0/7.0/**10.1** | 3.1/4.6/4.7/**6.9** |
| | Logit | 13.4/20.3/18.9/**22.1** | 2.7/3.8/5.3/**5.5** | 6.0/6.7/6.3/**8.0** | 16.0/20.2/**20.3**/19.7 | 9.5/12.8/12.7/**13.8** |
| | Margin | 16.5/17.3/21.7/**24.1** | 4.8/5.8/6.0/**6.4** | 7.6/**9.9**/9.0/9.7 | 19.5/22.5/23.2/**24.4** | 12.1/13.9/15.0/**16.2** |
| | SH | 17.1/**24.7**/22.4/23.6 | 3.7/3.4/5.8/**6.2** | 7.3/**9.2**/8.7/8.8 | 20.1/22.2/22.4/**23.9** | 12.1/14.9/14.8/**15.6** |
| | SU | 21.3/22.9/20.8/**24.2** | 5.0/4.1/5.0/**5.2** | 4.8/5.0/**6.8**/6.5 | 20.0/18.9/18.9/**20.6** | 12.8/12.7/12.9/**14.1** |
| Dense121 | CE | 1.7/3.6/4.3/**9.2** | 1.2/1.6/2.2/**3.4** | 1.2/2.2/2.3/**4.0** | 6.2/11.4/13.2/**16.9** | 2.6/4.7/5.5/**8.4** |
| | Logit | 10.5/12.4/12.6/**13.4** | 2.5/5.1/4.6/**5.2** | 4.7/6.5/6.6/**7.4** | 23.5/**29.4**/28.9/29.1 | 10.3/13.4/13.2/**13.8** |
| | Margin | 11.5/13.6/14.5/**16.1** | 3.6/5.3/6.4/**6.5** | 5.2/7.0/7.1/**8.0** | 20.8/28.0/28.8/**31.7** | 10.3/13.5/14.2/**15.6** |
| | SH | 9.3/14.4/15.7/**17.6** | 2.9/5.0/**5.2**/4.9 | 4.0/**7.2**/6.2/7.0 | 25.2/29.8/30.3/**31.9** | 10.4/14.1/14.4/**15.4** |
| | SU | 12.9/15.1/15.4/**16.2** | 4.4/4.4/4.5/**5.3** | 4.4/4.9/4.9/**5.1** | 23.9/**27.0**/25.2/26.3 | 11.4/12.9/12.5/**13.2** |
| VGG16 | CE | 0.0/0.4/0.5/**0.7** | 0.0/0.0/0.0/**0.2** | 0.0/0.0/0.1/**0.4** | 0.6/0.6/0.7/**2.2** | 0.2/0.3/0.3/**0.9** |
| | Logit | 6.2/7.6/8.6/**9.5** | 0.2/0.5/**0.8**/0.6 | 1.4/1.9/**3.0**/2.3 | 6.7/7.7/**7.9**/6.8 | 3.6/4.4/**5.1**/4.8 |
| | Margin | 6.4/7.4/7.8/**8.3** | 0.1/0.4/0.3/**0.5** | 1.8/1.8/2.6/**2.7** | 4.2/4.2/**6.2**/5.5 | 3.1/3.5/4.2/**4.3** |
| | SH | 7.1/8.7/9.2/**10.1** | 0.4/0.4/**1.0**/0.9 | 2.0/1.7/**2.9**/2.5 | 9.4/8.7/11.3/**11.5** | 4.7/4.9/6.1/**6.3** |
| | SU | 8.1/8.3/**9.9**/9.5 | 0.8/**0.9**/0.8/0.7 | 2.2/2.8/2.6/**2.9** | 12.7/11.8/11.4/**13.1** | 6.0/6.0/6.2/**6.6** |
| Inc-v3 | CE | 0.0/0.1/0.3/**0.7** | 0.2/0.2/0.2/**0.8** | 0.2/0.4/0.2/**0.8** | 0.7/0.8/0.6/**1.8** | 0.3/0.4/0.3/**1.0** |
| | Logit | 0.2/1.6/**1.8**/1.6 | 0.3/0.5/**1.2**/1.1 | 0.6/0.7/0.7/**0.9** | 1.0/1.0/**1.2**/1.1 | 0.4/1.0/**1.2**/1.2 |
| | Margin | 0.9/**1.4**/1.1/1.1 | 0.4/0.4/0.5/**0.7** | 0.4/0.5/0.6/**0.8** | 0.9/1.6/1.2/**1.6** | 0.7/1.0/0.9/**1.1** |
| | SH | 0.2/1.7/**1.9**/1.8 | 0.2/0.5/**0.8**/0.7 | 0.7/0.9/**1.0**/0.9 | 0.8/0.2/1.0/**1.2** | 0.5/0.8/**1.2**/1.2 |
| | SU | 0.9/1.1/1.2/**1.5** | 0.2/0.4/**0.6**/0.5 | 0.2/0.3/0.4/**0.6** | 1.0/0.9/0.9/**1.1** | 0.6/0.7/0.8/**0.9** |

**Table 2.** Targeted transfer success rate (%): no fine-tuning/ fine-tuning (ILA/FFT/AaF), in the most difficult-target scenario.

| Attack | Source Model: Res50 | | | Source Model: Dense121 | | |
|---|---|---|---|---|---|---|
| | →Inc-v3 | →Dense121 | →VGG16 | →Inc-v3 | →Res50 | →VGG16 |
| CE | 1.3/2.5/3.1/**9.8** | 25.8/44.8/45.3/**53.9** | 15.0/30.6/29.7/**41.6** | 1.2/4.5/6.1/**9.0** | 6.5/19.6/23.4/**34.8** | 3.6/14.7/19.2/**29.6** |
| Logit | 3.6/7.3/7.5/**10.2** | 51.6/56.6/53.1/**59.7** | 38.6/45.3/44.3/**46.6** | 3.5/7.6/8.3/**10.2** | 22.7/38.8/41.6/**45.1** | 18.3/31.5/37.5/**42.2** |
| Margin | 4.1/7.5/8.4/**8.6** | 52.7/60.7/57.1/**62.5** | 38.0/51.8/47.6/**53.1** | 4.0/9.5/8.7/**10.3** | 24.7/43.7/44.4/**49.2** | 18.2/36.9/35.0/**40.6** |
| SH | 4.0/7.0/8.1/**9.8** | 54.5/60.6/57.9/**63.3** | 41.6/49.7/51.2/**56.2** | 4.0/8.1/8.6/**10.5** | 24.5/41.9/43.3/**43.8** | 21.2/37.2/40.4/**41.1** |
| SU | 5.0/6.8/7.5/**12.3** | 56.2/55.9/54.8/**61.4** | 44.1/48.3/49.7/**53.6** | 4.4/7.6/8.0/**9.6** | 27.4/41.2/41.7/**46.2** | 24.3/37.7/37.6/**40.1** |

| Attack | Source Model: VGG16 | | | Source Model: Inc-v3 | | |
|---|---|---|---|---|---|---|
| | →Inc-v3 | →Res50 | →Dense121 | →Res50 | →Dense121 | →VGG16 |
| CE | 0.0/0.0/0.0/0.0 | 0.0/0.5/1.4/**1.8** | 0.0/0.4/0.6/**1.9** | 2.4/3.9/5.9/**7.8** | 4.2/4.9/7.8/**10.3** | 2.3/4.6/5.0/**8.1** |
| Logit | 0.3/0.4/0.4/**0.7** | 5.6/8.0/8.9/**12.6** | 7.0/8.8/8.5/**12.2** | 3.8/6.1/8.1/**10.4** | 5.5/7.2/10.5/**11.8** | 3.2/6.1/7.9/**8.4** |
| Margin | 0.0/0.3/0.4/**0.5** | 4.4/6.4/6.6/**9.2** | 6.0/8.6/**10.5**/10.5 | 2.5/5.2/7.5/**9.2** | 3.5/7.1/8.8/**12.4** | 2.0/4.3/6.2/**9.1** |
| SH | 0.1/0.2/0.3/**0.5** | 3.9/7.6/8.1/**9.5** | 6.8/8.6/9.4/**9.7** | 3.0/5.5/8.2/**10.1** | 4.9/7.4/10.6/**12.8** | 3.4/4.9/7.2/**10.0** |
| SU | 0.1/0.1/0.2/**0.8** | 5.7/7.1/7.6/**9.2** | 7.4/8.2/9.7/**10.4** | 3.9/5.8/8.1/**10.5** | 6.7/8.9/12.1/**14.3** | 3.9/5.3/9.6/**11.2** |

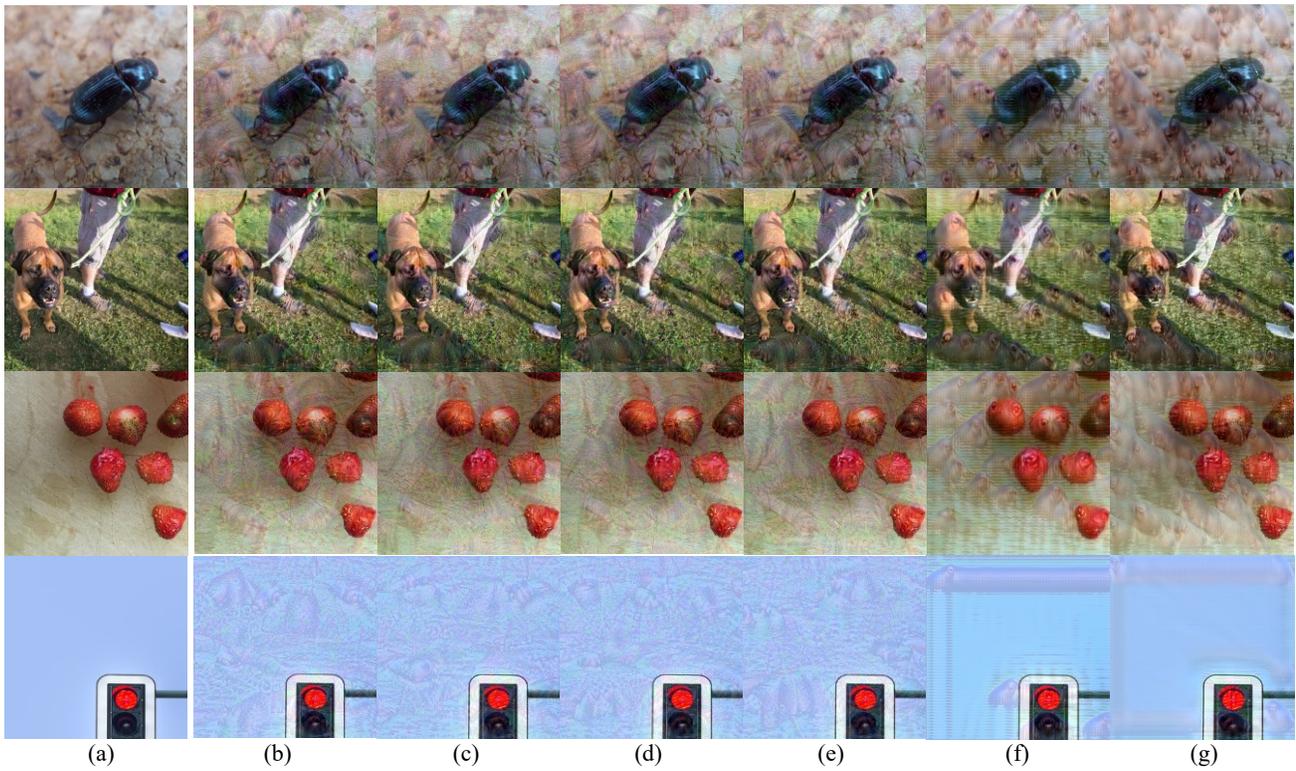

**Fig. 3.** The visual comparison of the AEs generated by different methods, $\epsilon = 16$. The target class is '*hippopotamus*'. (a) Original image, (b) Logit+*AaF*, (c) Margin+*AaF*, (d) SupHigh+*AaF*, (e) SU+*AaF*, (f) TTP, (g) C-GSP.